\documentclass[10pt,twocolumn,letterpaper]{article}

\usepackage{cvpr}              
\usepackage{graphicx}
\usepackage{caption} 
\usepackage{multirow}
\definecolor{cvprblue}{rgb}{0.21,0.49,0.74}
\usepackage[pagebackref,breaklinks,colorlinks,allcolors=cvprblue]{hyperref}

\def\confName{CVPR}
\def\confYear{2026}

\title{MVGGT: Multimodal Visual Geometry Grounded Transformer for \\ Multiview 3D Referring Expression Segmentation}
\author{
    Changli Wu$^{1,2,\dagger}$, 
    Haodong Wang$^{1,\dagger}$, 
    Jiayi Ji$^{1,*}$, 
    Yutian Yao$^{5}$, \\
    Chunsai Du$^{4}$, 
    Jihua Kang$^{4}$, 
    Yanwei Fu$^{2,3}$, 
    Liujuan Cao$^{1}$
    \\[0.2cm] 
    $^{1}$Key Laboratory of Multimedia Trusted Perception and Efficient Computing,\\
  Ministry of Education of China,
  Xiamen University \\
    $^{2}$Shanghai Innovation Institute \quad
    $^{3}$Fudan University \quad
    $^{4}$ByteDance \\
    $^{5}$Tianjin University of Science and Technology
    \\[0.2cm] 
    {\tt\small \{wuchangli, wahadon\}@stu.xmu.edu.cn, jjyxmu@gmail.com, 22201316@mail.tust.edu.cn,} \\ 
    {\tt\small \{duchunsai, kangjihua\}@bytedance.com, yanweifu@fudan.edu.cn, caoliujuan@xmu.edu.cn} 
}
\begin{document}
\twocolumn[{%
\renewcommand\twocolumn[1][]{#1}%
\maketitle 
\renewcommand{\thefootnote}{}  
\footnote{$^{\dagger}$Equal Contribution. $^{\ast}$Corresponding Author.}

\begin{center} 
\vspace{-.3cm}
\centering \includegraphics[width=0.98\textwidth]{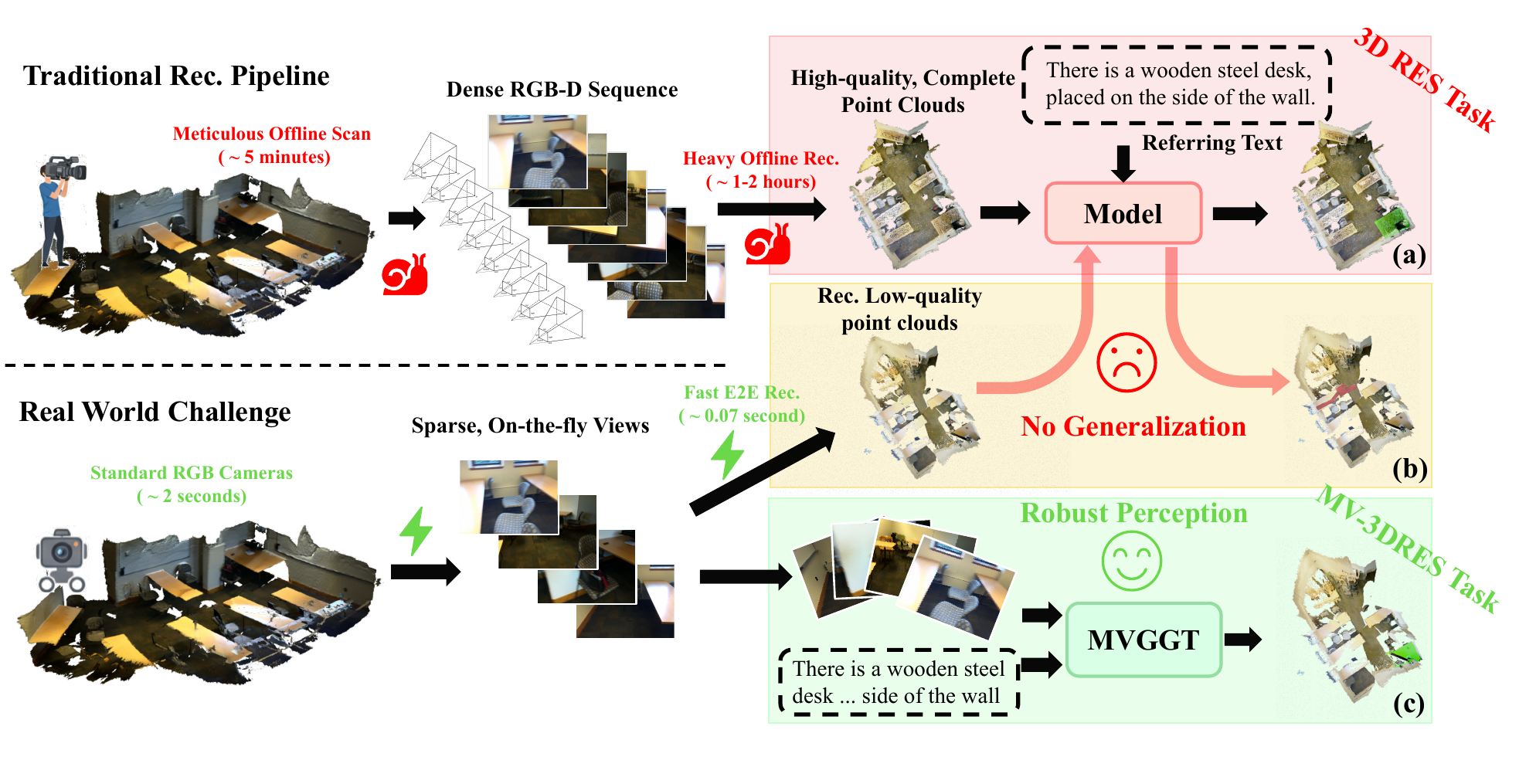} 
\vspace{-1.2em}
\captionof{figure}{\textbf{The Reality Gap: From Idealized 3D RES to Real-World MV-3DRES.} \textbf{(a)} Traditional 3D RES depends on dense, high-quality point clouds produced by slow offline scanning and heavy reconstruction.
\textbf{(b)} Applied to sparse, low-quality point clouds from real-world RGB views, these models fail to generalize.
\textbf{(c)} We introduce MV-3DRES, which uses sparse multi-view RGB inputs and text to achieve robust joint reconstruction and perception, enabled by our MVGGT model.} 
\label{fig:1}
\end{center}%
}]

{
  \renewcommand{\thefootnote}{}
  \footnotetext[1]{$^{\dagger}$Equal Contribution.}
  \footnotetext[2]{$^{*}$Corresponding Author.}
}

\begin{abstract}
Most existing 3D referring expression segmentation (3DRES) methods rely on dense, high-quality point clouds, while real-world agents such as robots and mobile phones operate with only a few sparse RGB views and strict latency constraints. We introduce Multi-view 3D Referring Expression Segmentation (MV-3DRES), where the model must recover scene structure and segment the referred object directly from sparse multi-view images. Traditional two-stage pipelines, which first reconstruct a point cloud and then perform segmentation, often yield low-quality geometry, produce coarse or degraded target regions, and run slowly. We propose the Multimodal Visual Geometry Grounded Transformer (MVGGT), an efficient end-to-end framework that integrates language information into sparse-view geometric reasoning through a dual-branch design. Training in this setting exposes a critical optimization barrier, termed Foreground Gradient Dilution (FGD), where sparse 3D signals lead to weak supervision. To resolve this, we introduce Per-view No-target Suppression Optimization (PVSO), which provides stronger and more balanced gradients across views, enabling stable and efficient learning. To support consistent evaluation, we build MVRefer, a benchmark that defines standardized settings and metrics for MV-3DRES. Experiments show that MVGGT establishes the first strong baseline and achieves both high accuracy and fast inference, outperforming existing alternatives. The code is available at https://mvggt.github.io/.
\end{abstract}
\section{Introduction}
\label{sec:introduction}
Grounding natural language in 3D physical scenes is fundamental to embodied AI. A prominent formulation is 3D referring expression segmentation (3DRES), where a model segments an object in a 3D scene given a textual description. Although recent methods~\cite{scanrefer,referit3d,multi3drefer,3drec_17,tgnn,3dstmn,3dgres,wu2024rg} have achieved strong results, they are built upon a rarely questioned assumption: the availability of dense, complete, and reliable point clouds. Such point clouds typically require LIDAR sensors or lengthy RGB-D SLAM pipelines like BundleFusion~\cite{dai2017bundlefusion}, which demand deliberate scanning and heavy offline processing. This assumption stands in stark contrast to real-world agents—robots, AR glasses, mobile devices—that perceive environments through only a few casually captured RGB views.

In real settings, high-fidelity geometry is the exception. Sparse multi-view images produce 3D reconstructions that are noisy, incomplete, and often ambiguous (Figure~\ref{fig:1}(b)). Existing 3DRES models, trained on idealized point clouds, collapse under such inputs. This exposes a fundamental limitation: current 3DRES is sensor-privileged and misaligned with the actual sensing conditions of embodied systems. It motivates a central question:
\textbf{How can we achieve language-grounded 3D perception when complete geometry is no longer given but must be inferred from sparse, inconsistent views?}

We address this by introducing Multi-view 3D Referring Expression Segmentation (MV-3DRES), a new setting where the model must jointly reconstruct the scene and segment the referred object directly from sparse RGB views (Figure~\ref{fig:1}(c)). MV-3DRES is inherently challenging: the model must reason over missing structure, integrate information across misaligned viewpoints, and resolve linguistic ambiguities without access to dense 3D input.

Conventional pipelines fail in this regime. Purely 2D methods cannot enforce global 3D consistency, since they operate on isolated views and cannot resolve depth ordering, occlusion relationships, or spatial relations such as ``in front of'' or ``on the left.'' As a result, back-projecting their per-view masks produces fragmented or conflicting 3D predictions. Two-stage ``reconstruct-then-segment'' pipelines face a different failure mode: sparse inputs yield point clouds that are noisy, incomplete, and structurally distorted, making it difficult for 3DRES models to recover full object extents. Moreover, running a full reconstruction before segmentation incurs substantial latency, limiting practical deployment.

To this end, we propose the Multimodal Visual Geometry Grounded Transformer (MVGGT), the first end-to-end architecture designed specifically for MV-3DRES. MVGGT adopts a dual-branch paradigm: a frozen geometric branch provides camera poses, depth cues, and a coarse structural scaffold, while a multimodal branch injects linguistic cues into sparse-view visual features through cross-view, cross-modal attention. This design embodies a key conceptual shift: language is intertwined with geometric reasoning from the start, enabling it to guide evidence aggregation and scene disambiguation long before a complete 3D representation exists.

\begin{figure}
    \centering
    \includegraphics[width=1.\linewidth]{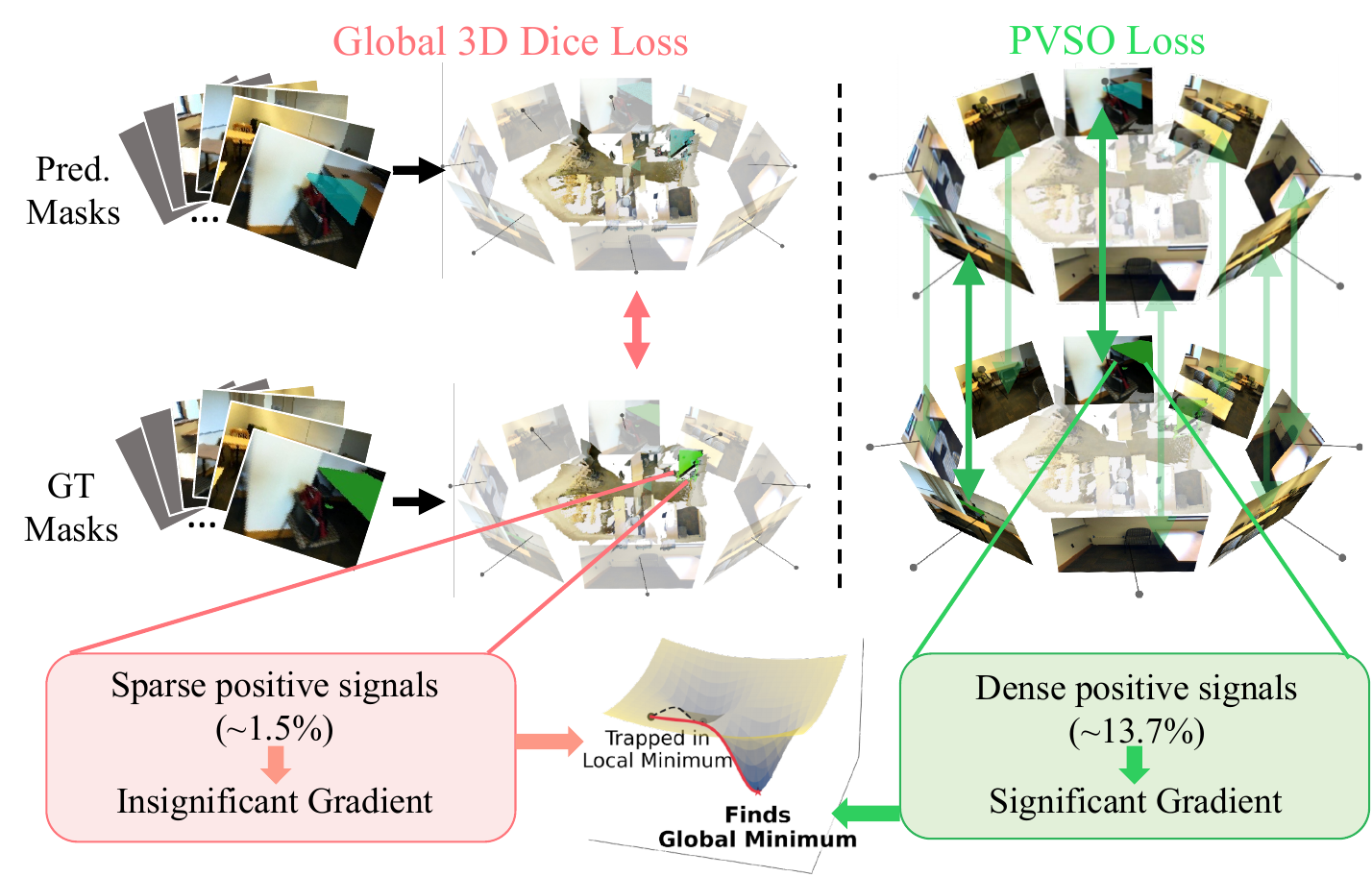}
    \caption{Illustration of Foreground Gradient Dilution Problem of Global 3D DICE loss and Per-view No-Target Suppression Optimization.}
\label{fig:2}
\end{figure}

However, sparse-view learning introduces a fundamental optimization challenge. Sparse multi-view reconstruction produces point clouds in which the target instance is represented by only a very small number of scattered points, far fewer than in the dense point clouds used by conventional 3DRES methods. Under such extreme foreground sparsity, standard 3D losses such as Dice become ineffective: gradients from the target region are overwhelmed by background points, leading to Foreground Gradient Dilution (FGD) and causing the optimization to stagnate in early training—gradients are too small to escape poor local minima, as illustrated in Figure~\ref{fig:2}. The problem is exacerbated by view-dependent visibility—some views contain clear target evidence, while others provide almost none—making uniform 3D supervision unstable and noisy. To mitigate FGD, we introduce Per-view No-target Suppression Optimization (PVSO), which shifts supervision back to 2D view space where the target occupies a larger and more reliable area. This per-view formulation amplifies meaningful gradients from informative views and suppresses misleading signals from target-absent views, resulting in significantly more stable and effective training.

Finally, to standardize evaluation, we construct MVRefer, the first benchmark defining settings, metrics, and data protocol for MV-3DRES. Extensive experiments show that MVGGT provides a strong baseline and significantly outperforms existing alternatives. Taken together, our contributions are fourfold:
\begin{itemize}
    \item We identify and formalize MV-3DRES, a new problem setting that aligns 3D grounding with realistic sensing conditions.
    \item We propose MVGGT, a novel dual-branch architecture unifying geometric scaffolding with cross-view, language-aware perception.
    \item We analyze and address the Foreground Gradient Dilution challenge via PVSO, offering a principled optimization strategy tailored for sparse 3D supervision.
    \item We construct the MVRefer benchmark, defining standardized settings and metrics for MV-3DRES and providing the first strong baseline.
\end{itemize}

\section{Related Work}
\subsection{Traditional 3D Referring Segmentation}
3D grounding~\cite{ jain2023oneformer, huang2022multi,3dvg1,3dvg2,3dvg3,3drec_2, 3dvg4,3drec_5,he2021transrefer3d,lai2022stratified,guo2023viewrefer,fang2024transcrib3d,liu2021refer} aims to locate a specific object in a 3D scene based on a unique natural language description~\cite{scanrefer, referit3d},which is part of vision-language tasks~\cite{v_l_3,v_l_4,v_l_5,v_l_6,v_l_7,liu2024primitivenet}. Following this, the 3D-RES (3D Referring Segmentation) task~\cite{liu2024weakly,deng20253d,huang2025point,chen20253drest,li2024laso,chen2024grounded,huang2025reason3d,huang2025mllm,hu2025omni} aims to segment a specific object within a point cloud based on a textual query. The field has evolved from foundational two-stage paradigms~\cite{tgnn,3drec_3,3dres_4} (relying on object proposals and language matching) to recent end-to-end architectures~\cite{3drec_5,3dres_1,3dres_2,3dres_3,3dres_4,3dgres,3dres_5,3dstmn,wu2024rg} demonstrating high efficacy through advanced cross-modal fusion. Despite this progress, the sub-field shares a critical limitation: the requirement for high-quality, dense point clouds. This expensive geometric data is inaccessible to many real-world embodied agents that must rely on sparse, online RGB captures.

\subsection{Multi-View Feed-forward Reconstruction}

Reconstructing 3D geometry from multi-view RGB images provides a practical solution to the input ambiguity in sparse-view settings. Early feed-forward approaches such as DUSt3R and MASt3R~\cite{leroy2024grounding,wang2024dust3r} introduced coupled scene representations but required heavy post-processing or integration with classical SfM/SLAM pipelines~\cite{duisterhof2025mast3r,elflein2025light3r,murai2025mast3r,pataki2025mp} for unconstrained reconstruction. Later works improved efficiency and stability by replacing classical optimization with Transformer-based latent state propagation, as demonstrated by Spann3R, CUT3R, and MUSt3R~\cite{wang20243d,wang2025continuous,cabon2025must3r}. Streaming models like WinT3R~\cite{wint3rwint3r} further enabled real-time performance via sliding-window processing and global camera token pooling.

More recent architectures—VGGT and its successors~\cite{wang2025vggt,pi3,shen2025fastvggt,wang2025faster,deng2025vggt}—adopt alternating-attention designs to achieve robust generalized reconstruction, while extensions address semantic and multi-task perception~\cite{wang2023mvtrans,transformeriggt,li2025ovseg3r,siu3r}. This progress culminates in universal backbones such as MapAnything~\cite{keetha2025mapanything}, which produce fully factored, metric-aware scene representations. 

\section{MV-3DRES Task and MVRefer Benchmark}
\label{sec:task_and_benchmark}

\subsection{Task Formulation}
\label{sec:formulation}

We formalize the Multi-view 3D Referring Segmentation (MV-3DRES) task to align 3D language grounding with the sensing constraints of real-world agents. Instead of assuming access to pre-constructed dense point clouds, the model operates directly on sparse multi-view RGB images.

Given a set of $N$ RGB views $I=\{I_i\}_{i=1}^{N}$ and a natural-language referring expression $T$, the goal is to learn a function
\begin{equation}
f: (I, T) \rightarrow (S', M),
\end{equation}
where $S' \in \mathbb{R}^{K \times 3}$ denotes the reconstructed 3D point cloud containing $K$ points, and $M \in \{0,1\}^K$ is the corresponding 3D binary mask marking the points that belong to the object referred to by $T$. The model must infer both geometry and semantics from the same sparse observations, without any ground-truth 3D input at inference time.

This formulation introduces challenges not present in standard 3DRES. Sparse multi-view observations generate incomplete and noisy geometry, forcing the model to couple reconstruction and grounding. Spatial relations described in language, such as “on the left of the chair,” must be resolved across viewpoints with inconsistent visibility. Moreover, the target object often occupies only a small portion of the available views, yielding severe foreground sparsity and weak supervisory signals, which we later characterize as Foreground Gradient Dilution (FGD).

\subsection{The MVRefer Benchmark}
\label{sec:benchmark}

To support systematic evaluation of MV-3DRES, we construct MVRefer, a benchmark built upon ScanRefer~\cite{scanrefer} and the underlying ScanNet sequences~\cite{ScanNet}. MVRefer is designed to emulate how an embodied agent perceives a scene through a limited number of casual views.

\subsubsection{Benchmark Setting}
\label{sec:benchmark_setting}

For each language–object pair in ScanRefer~\cite{scanrefer}, we sample $N = 8$ RGB frames from the raw ScanNet video stream~\cite{ScanNet} at uniform temporal intervals to approximate sparse, on-the-fly observations. Sparse sampling creates a solvability issue: the target may be absent from all selected frames. To ensure each sample remains resolvable, we perform a visibility validation step. If none of the initial eight images contain the target, we replace one no-target frame with a randomly chosen target-visible frame. This guarantees at least one positive view while naturally preserving a high proportion of no-target views, maintaining the difficulty inherent to sparse-view grounding.

\subsubsection{Evaluation Metrics and Splits}
\label{sec:benchmark_metrics}

Evaluating MV-3DRES requires metrics that disentangle grounding quality from reconstruction quality. Since both outputs in $(S', M)$ are jointly predicted from sparse inputs, reconstruction errors can obscure the model’s true grounding ability.

\paragraph{Traditional 3D Metric.}
We report global 3D mean IoU,
\begin{equation}
\text{mIoU}_{\text{global}}=\text{IoU}(M, M^{\ast}),
\end{equation}
where $M^{\ast}$ denotes the ground-truth mask projected onto the reconstructed point cloud $S'$. Although standard, $\text{mIoU}{\text{global}}$ entangles segmentation performance with the fidelity of $S'$, making it insufficient as the primary diagnostic measure.

\paragraph{Multi-view Diagnostic Metrics.}
To isolate grounding behaviors, we reproject the predicted 3D mask $M$ into each view using the known camera intrinsics and extrinsics. Let $P_i(M)$ denote the projected 2D mask for view $i$, and $P_i(M^{\ast})$ its ground-truth counterpart. We compute:
\begin{equation}
\text{mIoU}_{\text{view}} = \frac{1}{N} \sum_{i=1}^{N} \text{IoU}\left(P_i(M), P_i(M^{\ast})\right),
\end{equation}
\begin{equation}
\text{mIoU}_{\text{pos}} = \frac{1}{|\mathcal{V}^{+}|} \sum_{i \in \mathcal{V}^{+}} \text{IoU}\left(P_i(M), P_i(M^{\ast})\right),
\end{equation}
\begin{equation}
\text{mIoU}_{\text{neg}} = \frac{1}{|\mathcal{V}^{-}|} \sum_{i \in \mathcal{V}^{-}} \text{IoU}\left(P_i(M), P_i(M^{\ast})\right),
\end{equation}
where $\mathcal{V}^{+}$ and $\mathcal{V}^{-}$ denote the sets of target-visible and no-target views, respectively. These metrics shed light on grounding precision ($\text{mIoU}_{\text{pos}}$) and suppression ability ($\text{mIoU}_{\text{neg}}$), both of which are essential for robust performance under sparse supervision.

\paragraph{Difficulty Splits.}
To evaluate robustness under varying signal sparsity, we define two difficulty splits based on the target's 2D pixel ratio. A sample is categorized as \textit{hard} if the target occupies less than $5\%$ of pixels in all its visible views, and \textit{easy} if at least one view contains at least $5\%$ target pixels. This separation allows us to isolate performance differences arising from the strength of view-specific supervision.

\begin{figure*}
    \centering
    \includegraphics[width=.95\linewidth]{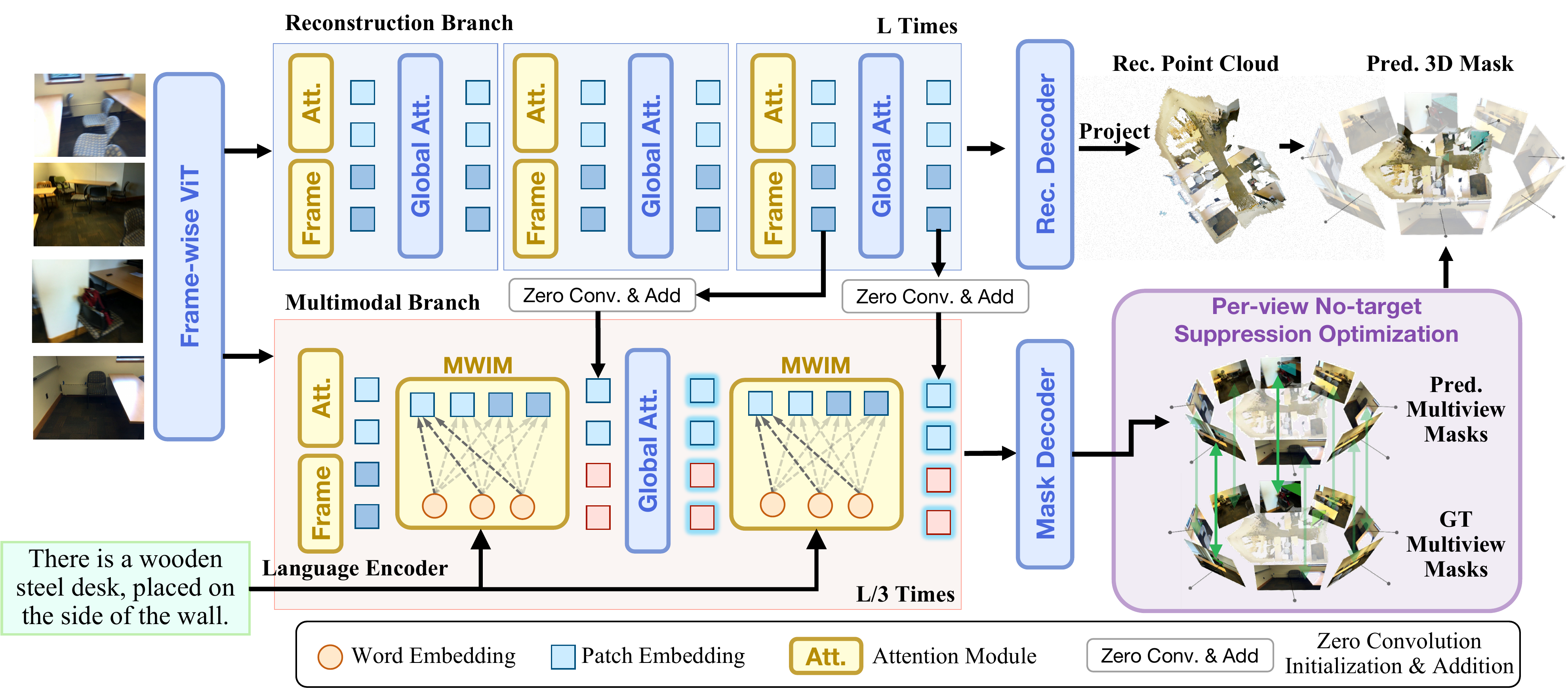}
    \caption{Architecture of MVGGT, which comprises a frozen Reconstruction Branch that establishes geometric structure and a trainable Multimodal Branch that integrates language into sparse-view visual reasoning.}
\label{fig:3}
\vspace{-.3cm}
\end{figure*}

\section{Method}
\label{sec:methodology}

We introduce the \textbf{Multimodal Visual Geometry Grounded Transformer (MVGGT)}, an end-to-end dual-branch framework tailored for the MV-3DRES task, as shown in Figure~\ref{fig:3}.

\subsection{The proposed MVGGT}
\label{sec:architecture}

MVGGT is designed to jointly recover 3D geometry and perform language-conditioned segmentation from sparse views. The separation into two branches allows the model to exploit a stable geometric scaffold while learning multimodal representations aligned with the text query.

\paragraph{Inputs and Encoders.}
Given $N$ input images $I=\{I_i\}_{i=1}^N$ and a referring expression $T$, each image is encoded by a frame-wise ViT, yielding patch embeddings $F_i^{\text{vis}}\in\mathbb{R}^{P\times D}$, where $P$ is the number of patches and $D$ the feature dimension. The text is tokenized and processed by a language encoder to produce word embeddings $F^{\text{lang}}\in\mathbb{R}^{W\times D}$, where $W$ denotes token count.

\paragraph{Frozen Reconstruction Branch.}
The reconstruction branch is a geometry-aware transformer with $L$ blocks. Each block alternates between frame-level self-attention and global cross-view attention, progressively building view-consistent structural cues. Let $F_{\ell}^{\text{geo}}$ denote the features at block $\ell\in\{1,\dots,L\}$. These features are fed to a reconstruction decoder that predicts camera poses and depth maps, which are back-projected into a coarse point cloud $S'$. All parameters in this branch remain frozen, ensuring a stable geometric prior across training and removing the need to re-learn 3D geometry from sparse images.

\paragraph{Trainable Multimodal Branch.}
The multimodal branch contains $L_{\text{multi}}=L/3$ transformer blocks. Its goal is to fuse geometric cues with text-conditioned visual features. Since the two branches have different depths, we align their interactions by injecting geometric features from the final $L/3$ blocks of the reconstruction branch into all $L_{\text{multi}}$ blocks of the multimodal branch.

\emph{Geometric Injection.}
The multimodal branch contains $L_{\text{multi}}=L/3$ blocks, and each of them receives geometric guidance from the \emph{final} $L_{\text{multi}}$ blocks of the reconstruction branch. Concretely, for the $l'$-th multimodal block ($l'=1,\dots,L_{\text{multi}}$), we take the geometric feature $F_{l}^{\text{geo}}$ from the $l$-th reconstruction block, where $l$ runs over the last $L_{\text{multi}}$ layers in order (i.e., the reconstruction branch’s $(L-L_{\text{multi}}+1)$-th layer provides geometry to the first multimodal block, the next layer to the second, and so on).

These geometric features are passed through a zero-initialized $1\times1$ convolution $\mathcal{Z}$~\cite{controllnet}, which projects them into the multimodal feature space. The input to the $l'$-th multimodal block is then
\begin{equation}
F_{l'}^{\text{in}}
  = F_{l'-1}^{\text{out}} + \mathcal{Z}(F_{l}^{\text{geo}}),
\end{equation}
where $F_{l'-1}^{\text{out}}$ is the output of the preceding visual attention and cross attention.  It allows the multimodal branch to progressively incorporate higher-level geometric cues without perturbing the pretrained reconstruction backbone. $F_{l'}^{\text{in}}$ is then fed into the visual attention module to get $F_{l'}^{\text{vis}}$.

\emph{Language Injection.}  
Within each multimodal block, language injection is implemented through a standard cross-attention layer. Let $F_{l'}^{\text{in}}$ be the visual tokens entering block $l'$. Query, key, and value projections are defined as
\begin{equation}
Q = F_{l'}^{\text{vis}} W_Q,\quad
K = F^{\text{lang}} W_K,\quad
V = F^{\text{lang}} W_V,
\end{equation}
with $W_Q$, $W_K$, and $W_V\in\mathbb{R}^{D\times D}$ learnable matrices. Attention is computed as
\begin{equation}
F_{l'}^{\text{out}}
    = \mathrm{softmax}\!\left(\frac{QK^\top}{\sqrt{D}}\right)V,
\end{equation}
and added back through a residual path. This multi-level injection allows text cues to guide feature aggregation across views.

\paragraph{Decoders and Outputs.}
Finally, the multimodal features $F_{L_{\text{multi}}}^{\text{multi}}$ are decoded into per-view masks $\{M_i\}_{i=1}^N$. Using the depths and camera parameters from the frozen reconstruction branch, these 2D masks are back-projected and aggregated on the reconstructed point cloud $S'$ to obtain the final 3D mask $M$. The resulting multi-view predictions serve as the supervision targets in PVSO (Section~\ref{subsec:pvso}).

\subsection{Foreground Gradient Dilution}
\label{subsec:fgd}

Training under the MV-3DRES setting is fundamentally hindered by the extreme sparsity of foreground points in the reconstructed 3D space. Let the Dice loss for 3D segmentation be
\begin{equation}
\mathcal{L}_{\text{Dice}}
= 1 - \frac{2I}{U},
\quad
I=\sum_{j} p_j g_j,
\quad
U=\sum_j p_j + \sum_j g_j,
\end{equation}
where $p_j$ is the predicted probability at point $j$ and $g_j\in\{0,1\}$ denotes ground-truth labels. The gradient with respect to $p_j$ is
\begin{equation}
\frac{\partial \mathcal{L}_{\text{Dice}}}{\partial p_j}
= \frac{2(I - g_j U)}{U^2}.
\end{equation}

During early training, predictions remain small and diffuse, yielding $I\approx0$. In MV-3DRES, reconstructed point clouds from sparse views are large (often $10^6$–$10^7$ points) while the target instance typically occupies less than $2\%$ of them. Consequently, the union term $U$ is dominated by background, inflating by several orders of magnitude. For a foreground point ($g_j=1$),
\begin{equation}
\left.\frac{\partial \mathcal{L}_{\text{Dice}}}{\partial p_j}\right|
_{g_j=1}
\approx -\frac{2}{U},
\end{equation}
whose magnitude becomes extremely small when $U$ is large. Empirically, gradients fall to $10^{-9}$–$10^{-11}$, far below the scale needed to drive meaningful updates. Although foreground points are present, their contribution to optimization becomes negligible, leading to stalled convergence. We refer to this failure mode as \emph{Foreground Gradient Dilution (FGD)}.

\subsection{Per-view No-Target Suppression Optimization}
\label{subsec:pvso}

To mitigate FGD, we propose \textbf{Per-view No-Target Suppression Optimization (PVSO)}, a view-wise supervision strategy that shifts early learning signals from sparse 3D space to the denser 2D image domain. This modification significantly reduces the imbalance between foreground and background, thereby amplifying effective gradients.

\paragraph{Positive-aware Sampling.}
Given the view set $\mathcal{V}$ of a scene, let $\mathcal{V}_t$ and $\mathcal{V}_n$ denote target-visible and no-target views, respectively. PVSO samples a subset $\mathcal{V}'$ while enforcing a minimum foreground-view ratio
\begin{equation}
\rho_t = \frac{|\mathcal{V}_t|}{|\mathcal{V}'|},
\end{equation}
ensuring that each batch contains sufficient positive evidence. This prevents the optimization from collapsing to trivial background predictions.

\paragraph{2D Gradient Concentration.}
For each sampled view, the predicted 3D mask is projected onto the image plane, and a 2D Dice loss is applied. Since foreground regions typically occupy $10$–$15\%$ of pixels in visible views—far larger than their $<2\%$ proportion in 3D point clouds—the 2D Dice denominator $U_{\text{2D}}$ becomes substantially smaller:
\begin{equation}
U_{\text{2D}} \ll U_{\text{3D}}.
\end{equation}
Given that the Dice gradient scales as $\mathcal{O}(1/U)$, the per-view supervision yields foreground gradients that are 1–3 orders of magnitude larger than in 3D. This concentrated 2D supervision strengthens early training signals.

\paragraph{Suppression of No-Target Views.}
No-target views often far outnumber target-visible ones. To avoid overwhelming the loss with trivial negatives, PVSO normalizes their contribution using
\begin{equation}
w_s = \frac{1}{|\mathcal{V}_n|}.
\end{equation}
The complete PVSO objective is
\begin{equation}
\begin{split}
L_{\text{PVSO}}
  = \frac{1}{|\mathcal{V}_t| + 1}
     \big(
        &\sum_{i\in\mathcal{V}_t}
            L_{\text{Dice}}(m_i, M_i^{\text{gt}})
        + \\
        &w_s
        \sum_{j\in\mathcal{V}_n}
            L_{\text{Dice}}(m_j, \mathbf{0})
     \big),
\end{split}
\end{equation}
where $m_i$ is the predicted 2D mask for view $i$ and $M_i^{\text{gt}}$ is the corresponding ground-truth mask (empty for no-target views).

\paragraph{Joint Objective.}
PVSO complements 3D supervision by providing dense, stable gradients during early training. The complete objective is
\begin{equation}
L_{\text{total}} = L_{\text{BCE}} + \lambda_p L_{\text{PVSO}},
\end{equation}
where $\lambda_p$ balances 2D and 3D signals. This formulation alleviates foreground gradient dilution and yields robust multimodal 3D grounding from sparse views.
\begin{table*}[!t]
\footnotesize
\centering
\caption{Performance comparison on the MVRefer benchmark.Metrics are all mIoU under different categories.}
\setlength{\tabcolsep}{3pt}
\resizebox{1\textwidth}{!}{
\begin{tabular}{l|cccc|cccc|cccc|cccc|cccc}
\toprule
\multirow{2}{*}{Method}& \multicolumn{4}{c|}{\textbf{Hard ($\sim$40\%)}} 
& \multicolumn{4}{c|}{\textbf{Easy ($\sim$60\%)}} 
& \multicolumn{4}{c|}{\textbf{Unique ($\sim$19\%)}} 
& \multicolumn{4}{c|}{\textbf{Multiple ($\sim$81\%)}} 
& \multicolumn{4}{c}{\textbf{Overall}} \\
& {\text{global}} & {\text{view}} & {\text{pos}} & {\text{neg}} 
& {\text{global}} & {\text{view}} & {\text{pos}} & {\text{neg}} 
& {\text{global}} & {\text{view}} & {\text{pos}} & {\text{neg}} 
& {\text{global}} & {\text{view}} & {\text{pos}} & {\text{neg}} 
& {\text{global}} & {\text{view}} & {\text{pos}} & {\text{neg}} \\
\midrule
two-stage
& 8.1 & 8.6 & 22.6 & 10.0
& 25.8 & 28.2 & 45.6 & 21.6
& 27.5 & 41.1 & 52.1 & 30.1
& 16.4 & 15.4 & 32.1 & 13.8
& 18.5 & 20.3 & 35.9 & 16.9
\\
2D-Lift 
& 6.4 & 15.0 & 25.3 & 11.9 
& 25.4 & 24.1 & 45.1 & 12.3 
& 31.5 & 30.9 & 47.2 & 19.2 
& 14.5 & 17.9 & 34.8 & 10.4 
& 17.8 & 20.4 & 37.2 & 12.1
\\
\textbf{Ours} 
& \textbf{24.4} & \textbf{67.3} & \textbf{31.6} & \textbf{78.0} 
& \textbf{50.1} & \textbf{70.6} & \textbf{52.2} & \textbf{81.2} 
& \textbf{65.2} & \textbf{82.6} & \textbf{64.5} & \textbf{90.4} 
& \textbf{33.8} & \textbf{66.1} & \textbf{39.0} & \textbf{77.4} 
& \textbf{39.9} & \textbf{69.3} & \textbf{44.0} & \textbf{79.9} \\
\bottomrule
\end{tabular}}
\label{tab:mv3dres}
\end{table*}

\begin{table}[!t]
\footnotesize
\centering
\caption{The results of traditional 3D-RES and MV-3DRES tasks under original ScanRefer setting.}
\setlength{\tabcolsep}{1pt}
\resizebox{1\linewidth}{!}{
\begin{tabular}{l|ccc|ccc|ccc}
\toprule
\multirow{2}{*}{Method}& \multicolumn{3}{c|}{\textbf{Unique ($\sim$19\%)}} 
& \multicolumn{3}{c|}{\textbf{Multiple ($\sim$81\%)}} 
& \multicolumn{3}{c}{\textbf{Overall}} \\
&{\text{Acc@25}} & {\text{Acc@50}} & {\text{mIoU}}
& {\text{Acc@25}} & {\text{Acc@50}} & {\text{mIoU}}
& {\text{Acc@25}} & {\text{Acc@50}} & {\text{mIoU}} \\
\midrule
\midrule
\multicolumn{10}{c}{\textit{Traditional 3D-RES (with ground-truth point clouds)}} \\
\midrule
TGNN~\cite{tgnn}
& 69.3 &  57.8 & 50.7 & 31.2 & 26.6 & 23.6 & 38.6 & 32.7 & 28.8
\\
3D-STMN~\cite{3dstmn}
& 89.3 & 84.0 & 74.5 & 46.2 & 29.2 & 31.1 & 54.6 &39.8 & 39.5
\\
SegPoint~\cite{3dres_3}
& - & - & - & - & - & - & - & - & 41.7
\\
Reason3D~\cite{huang2025reason3d}
& 88.4 & 84.2 & 74.6 & 50.5 & 31.7 & 34.1 & 57.9 & 41.9 & 42.0
\\
RG-SAN~\cite{wu2024rg}
& 89.2 & 84.3 & 74.5 & 55.0 & 35.4 & 37.4 & 61.7 & 44.9 & 44.6
\\
LESS~\cite{3dres_5}
& - & - & - & - & - & - & 53.2 & 29.9 & 33.7
\\
3D-LLaVA~\cite{deng20253d}
& - & - & - & - & - & - & - & - & 43.3
\\
\midrule 
\midrule
\multicolumn{10}{c}{\textit{MV-3DRES (sparse RGB inputs only)}} \\
\midrule
two-stage
& 43.8 & 20.9 & 27.5 & 25.1 & 12.2 & 16.4 & 28.7 & 13.9 & 18.5
\\
2D-Lift
& 51.6 & 26.0 & 31.5 & 21.5 & 5.6 & 14.5 & 27.3 & 9.6 & 17.8
\\
\textbf{Ours}
& \textbf{83.6} & \textbf{74.5} & \textbf{65.2} & \textbf{49.2} & \textbf{33.5} & \textbf{33.8} & \textbf{55.9} & \textbf{41.5} & \textbf{39.9}
\\
\bottomrule
\end{tabular}
}
\label{tab:scanrefer}
\end{table}

\section{Experiments}
\label{sec:experiments}

\subsection{Implementation Details and Setup}
\label{sec:setup}

\paragraph{MVGGT Configuration.}
We adopt the Pi3~\cite{pi3} reconstruction backbone (36 blocks), which remains frozen throughout training. We use a frozen Roberta model~\cite{liu2019roberta} as the language encoder. The multimodal branch contains $L_{\text{multi}}=12$ blocks and is optimized end-to-end. Training uses AdamW~\cite{loshchilov2017decoupled} with a $1\!\times\!10^{-4}$ learning rate, batch size of 16, and 30 epochs on a single NVIDIA 4090 GPU. The PVSO weight $\lambda_p$ is fixed to $1$.

\paragraph{Dataset and Metrics.}
All evaluations are conducted on the proposed \textbf{MVRefer} benchmark. We follow standard ScanRefer train/validation splits~\cite{scanrefer}. Alongside $\text{mIoU}_{\text{global}}$, we report all diagnostic view-level metrics from Section~\ref{sec:benchmark_metrics}, with special focus on the \textbf{Hard} and \textbf{Easy} subsets, which most directly reflect the foreground sparsity challenge underlying FGD.

\subsection{Main Results}
We evaluate MVGGT against two MV-3DRES baselines: (1) \textbf{2D-Lift}, which projects ReferDINO~\cite{referdino} masks into 3D via Pi3~\cite{pi3}; and (2) \textbf{two-stage}, a sequential pipeline of Pi3~\cite{pi3} reconstruction and LESS~\cite{3dres_5} segmentation.
As shown in Table~\ref{tab:mv3dres}, MVGGT consistently leads across all difficulty levels. On \textbf{Hard}, it achieves 24.4 global mIoU, surpassing two-stage and 2D-Lift by 16.3 and 18.0, respectively, with a 52.3 gain in view mIoU. This pattern holds for \textbf{Easy} (50.1 global; 70.6 view mIoU). Overall, MVGGT attains 39.9 global mIoU, exceeding the strongest baseline by 22.1. These gains validate MVGGT’s robustness to sparsity and the effectiveness of PVSO in mitigating FGD.


Table~\ref{tab:scanrefer} evaluates MVGGT with traditional 3D-RES and recent MV-3DRES baselines under the standard ScanRefer protocol. Despite using only sparse RGB, MVGGT achieves 65.2 mIoU on \emph{Unique}, significantly closing the gap with full 3D-resourced methods. In \emph{Multiple}, MVGGT reaches 33.8 mIoU, outperforming two-stage and 2D-Lift by 17.4 and 19.3. These results demonstrate that MVGGT enables reliable 3D grounding without ground-truth point clouds.

\subsection{Ablation Studies}
\label{sec:ablations}

We conduct comprehensive ablation studies to validate the effectiveness of each proposed component. All experiments are performed on the MVRefer benchmark.

\paragraph{Impact of Core Components.}

Table~\ref{tab:ablation_components} evaluates core components on the MVRefer benchmark. Removing both MVGGT and PVSO causes a significant performance drop, confirming the difficulty of sparse-view 3D grounding. Individually, PVSO improves $\mathrm{mIoU}_{\text{global}}$ ($26.9 \to 32.0$) and $\mathrm{mIoU}_{\text{view}}$ ($41.1 \to 47.5$) by mitigating Foreground Gradient Dilution. MVGGT enhances robustness, particularly on hard scenes ($19.0 \to 24.4$ $\mathrm{mIoU}_{\text{global}}$), by guiding geometric reasoning with linguistic cues. The full model achieves $39.9$ $\mathrm{mIoU}_{\text{global}}$ and $69.3$ $\mathrm{mIoU}_{\text{view}}$, validating the synergy between our multimodal design and optimization strategy.

\begin{table}[h]
\centering
\caption{Ablation studies on the core components.}
\label{tab:ablation_components}
\setlength{\tabcolsep}{3.5pt}
\resizebox{1\linewidth}{!}{
\begin{tabular}{cc|cc|cc|cc}
\toprule
 &  & \multicolumn{2}{c|}{\textbf{Easy $\uparrow$}} & \multicolumn{2}{c|}{\textbf{Hard $\uparrow$}} & \multicolumn{2}{c}{\textbf{Overall $\uparrow$}} \\
 \multirow{-2}{*}{\textbf{PVSO}}& \multirow{-2}{*}{\textbf{MVGGT}}& $\mathrm{mIoU}_{\text{global}}$ & $\mathrm{mIoU}_{\text{view}}$ & $\mathrm{mIoU}_{\text{global}}$ & $\mathrm{mIoU}_{\text{view}}$ & $\mathrm{mIoU}_{\text{global}}$ & $\mathrm{mIoU}_{\text{view}}$ \\
\midrule
\multicolumn{2}{c|}{2D-Lift} & 25.4 & 24.1 & 6.4 & 15.0 & 17.8 & 20.4 \\
\midrule
$\times$ & $\times$ & 36.3 & 43.0 & 12.9 & 38.5 & 26.9 & 41.1 \\
$\checkmark$ & $\times$ & 40.7 & 48.9 & 19.0 & 45.4 & 32.0 & 47.5 \\
$\checkmark$ & $\checkmark$ & \textbf{50.1} & \textbf{70.6} & \textbf{24.4} & \textbf{67.3} & \textbf{39.9} & \textbf{69.3} \\
\bottomrule
\end{tabular}
}
\end{table}

\begin{figure*}[t]
    \centering
    \includegraphics[width=1.\textwidth]{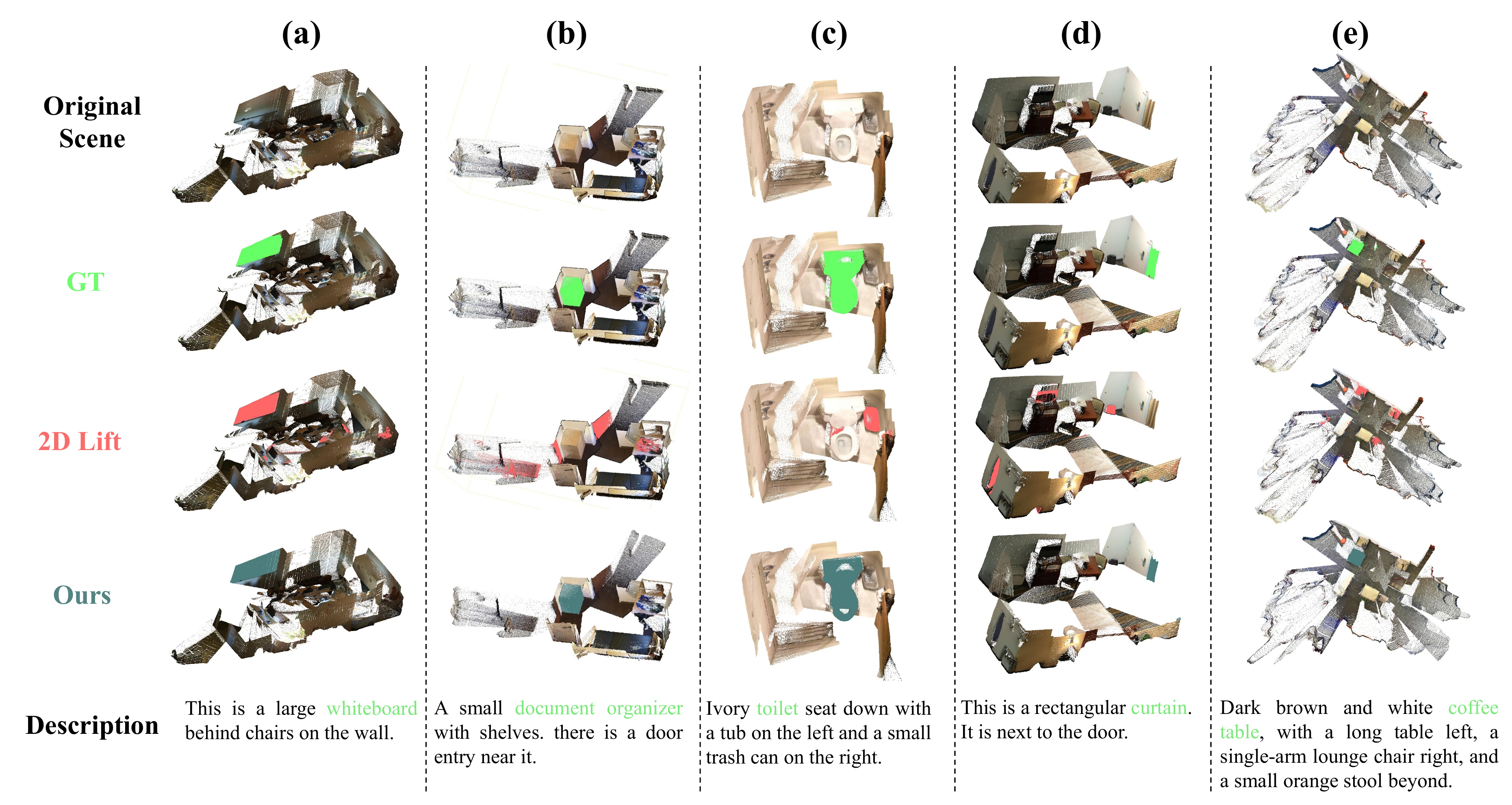} 
    \caption{
        Qualitative comparison on the MVRefer benchmark.
    }
    \label{fig:qualitative}
\end{figure*}

\paragraph{PVSO Component Analysis.}
Table~\ref{tab:ablation_pvso} evaluates PVSO variants. Without no-target suppression, random sampling yields unstable performance. Enabling suppression improves $\mathrm{mIoU}_{\text{global}}$ ($32.4 \to 36.7$) by filtering misleading gradients from target-absent views, while hybrid sampling ensures sufficient per-batch positive evidence. A $0.5$ no-target ratio achieves the best balance ($39.9$ $\mathrm{mIoU}_{\text{global}}$, $69.3$ $\mathrm{mIoU}_{\text{view}}$); extreme ratios either underexploit negative pairs or drown out positives. These results confirm that stable sparse-view learning requires weighting view contributions by their actual target visibility.

\begin{table}[h]
\centering
\caption{Ablation on PVSO components.}
\label{tab:ablation_pvso}
\setlength{\tabcolsep}{1.5pt}
\resizebox{1\linewidth}{!}{
\begin{tabular}{lcc|cc|cc|cc}
\toprule
\textbf{Sampling} & \textbf{No-Target} & \textbf{No-Target} & \multicolumn{2}{c|}{\textbf{Easy $\uparrow$}} & \multicolumn{2}{c|}{\textbf{Hard $\uparrow$}} & \multicolumn{2}{c}{\textbf{Overall $\uparrow$}} \\
\textbf{Strategy}& \textbf{Ratio} & \textbf{Suppression}& $\mathrm{mIoU}_{\text{global}}$ & $\mathrm{mIoU}_{\text{view}}$ & $\mathrm{mIoU}_{\text{global}}$ & $\mathrm{mIoU}_{\text{view}}$ & $\mathrm{mIoU}_{\text{global}}$ & $\mathrm{mIoU}_{\text{view}}$ \\
\midrule
Random & - & $\times$ & 40.8 & 65.1 & 19.7 & 66.5 & 32.4 & 65.6 \\
Random & - & \checkmark & 46.2 & 55.8 & 22.4 & 49.3 & 36.7 & 53.2 \\
Hybrid & 0 & \checkmark & 42.4 & 24.3 & 11.2 & 10.3 & 29.9 & 18.7 \\
Hybrid & 0.25 & \checkmark & 46.5 & 65.2 & 19.4 & 59.6 & 35.7 & 63.0 \\
Hybrid & 0.5 & \checkmark & \textbf{50.1} & \textbf{70.6} & \textbf{24.4} & \textbf{67.3} & \textbf{39.9} & \textbf{69.3} \\
Hybrid & 0.75 & \checkmark & 30.3 & 64.0 & 18.7 & 72.0 & 25.7 & 67.2 \\
\bottomrule
\end{tabular}
}
\end{table}

\paragraph{MVGGT Fusion Architecture.}
Table~\ref{tab:ablation_fusion} evaluates multimodal fusion placement. Early fusion performs weakest, as injecting language before geometric features form can disrupt structural reasoning. Middle fusion shows modest gains, while late fusion yields the best results ($39.9$ $\mathrm{mIoU}_{\text{global}}$; $69.3$ $\mathrm{mIoU}_{\text{view}}$). This trend suggests that spatial perception should be established first, with language providing later refinement. Such an order ensures more stable and discriminative cross-view representations.

\begin{table}[h]
\centering
\caption{Ablation on MVGGT fusion stage.}
\label{tab:ablation_fusion}
\setlength{\tabcolsep}{3.5pt}
\resizebox{1\linewidth}{!}{
\begin{tabular}{l|cc|cc|cc}
\toprule
\multirow{2}{*}{\textbf{Fusion Stage}} & \multicolumn{2}{c|}{\textbf{Easy $\uparrow$}} & \multicolumn{2}{c|}{\textbf{Hard $\uparrow$}} & \multicolumn{2}{c}{\textbf{Overall $\uparrow$}} \\
 & $\mathrm{mIoU}_{\text{global}}$ & $\mathrm{mIoU}_{\text{view}}$ & $\mathrm{mIoU}_{\text{global}}$ & $\mathrm{mIoU}_{\text{view}}$ & $\mathrm{mIoU}_{\text{global}}$ & $\mathrm{mIoU}_{\text{view}}$ \\
\midrule
Early & 45.7 & 65.9 & 21.6 & 63.4 & 36.1 & 64.9 \\
Middle & 47.5 & 65.7 & 22.5 & 62.1 & 37.5 & 64.3 \\
Late & \textbf{50.1} & \textbf{70.6} & \textbf{24.4} & \textbf{67.3} & \textbf{39.9} & \textbf{69.3} \\
\bottomrule
\end{tabular}
}
\end{table}

\paragraph{Multimodal Branch Depth.}
Table~\ref{tab:ablation_layers} evaluates branch depth. A 6-layer configuration lacks alignment capacity, while 12 layers yield peak performance. Expanding to 16 layers causes a sharp decline due to overfitting and unstable attention under sparse supervision. These results suggest that moderate depth optimally balances expressiveness and regularity, ensuring language effectively guides aggregation without overwhelming the geometric signal.

\begin{table}[h]
\centering
\caption{Ablation on layer number of multimodal branch.}
\label{tab:ablation_layers}
\setlength{\tabcolsep}{3.5pt}
\resizebox{1\linewidth}{!}{
\begin{tabular}{c|cc|cc|cc}
\toprule
\multirow{2}{*}{\textbf{Layers}} & \multicolumn{2}{c|}{\textbf{Easy $\uparrow$}} & \multicolumn{2}{c|}{\textbf{Hard $\uparrow$}} & \multicolumn{2}{c}{\textbf{Overall $\uparrow$}} \\
 & $\mathrm{mIoU}_{\text{global}}$ & $\mathrm{mIoU}_{\text{view}}$ & $\mathrm{mIoU}_{\text{global}}$ & $\mathrm{mIoU}_{\text{view}}$ & $\mathrm{mIoU}_{\text{global}}$ & $\mathrm{mIoU}_{\text{view}}$ \\
\midrule
6 & 47.1 & 66.7 & 22.5 & 64.1 & 37.3 & 65.7 \\
12 & \textbf{50.1} & \textbf{70.6} & \textbf{24.4} & \textbf{67.3} & \textbf{39.9} & \textbf{69.3} \\
16 & 35.6 & 26.2 & 10.1 & 14.2 & 25.4 & 21.4 \\
\bottomrule
\end{tabular}
}
\end{table}

\subsection{Qualitative Analysis}
\label{sec:qualitative}

Fig.~\ref{fig:3} demonstrates MVGGT’s robustness under sparse and noisy views. Unlike 2D-lifting baselines that often drift or collapse under occlusion and depth ambiguity, MVGGT maintains stable, target-aligned segmentations. Specifically, it leverages geometric-linguistic fusion to resolve planar ambiguities (a) and semantic distraction (c). PVSO-balanced supervision further enables precise localization in clutter (b), while the architecture ensures fine-grained discrimination (d) and cross-view consistency despite partial visibility (e).

\section{Conclusion}

We introduce MV-3DRES, a sparse-view 3D grounding setting reflecting real-world agent conditions. To overcome the failure of conventional two-stage pipelines under sparsity, we propose MVGGT, a dual-branch architecture that integrates linguistic cues directly into geometric reasoning. We further address the Foreground Gradient Dilution challenge via PVSO, enabling stable optimization. Together with the MVRefer benchmark, our framework provides a practical path for robust multimodal 3D grounding and embodied perception.

\section*{Acknowledgements}
This work is supported by the National Key Research and Development Program of China (No. 2025YFE0113500), National Science Fund for Distinguished Young Scholars (No.62525605) and the National Natural Science Foundation of China (No. U25B2066, No. U22B2051, No. 62302411).
{
    \small
    \bibliographystyle{ieeenat_fullname}
    \bibliography{main}
}

\end{document}